\documentclass{new_tlp}

\usepackage{url}
\usepackage[T1]{fontenc}
\usepackage{times}
\usepackage{graphicx}

\renewcommand{\figurename}{Table}

\newcommand{\K}{\mbox{\rm K}}
\newcommand{\M}{\mbox{\rm M}}
\newcommand{\Kell}{\mbox{\ensuremath{\rm K\hspace{0.05cm}\ell}}}
\newcommand{\Mell}{\mbox{\ensuremath{\rm M\hspace{0.05cm}\ell}}}
\newcommand{\ASPor}{\mbox{ ~{\tt or}~ }}
\newcommand{\ASPnot}{\mbox{ {\tt not} }}
\newcommand{\leftASPnot}{\mbox{{\tt not} }}
\newcommand{\ssASPnot}{\mbox{\hspace{0.05cm}{\tt not}\hspace{0.05cm}}}
\newcommand{\ssleftASPnot}{\mbox{{\tt not}\hspace{0.05cm}}}

\newcommand{\true}{\mbox{\sf{\scalebox{0.85}[0.85]{\emph{true}}}}}

\newcommand\mystrut{\rule{0pt}{7pt}}
\newcommand{\WVC}{{%
  \setbox0\hbox{\ensuremath{\leftarrow}}%
  \rlap{\hbox to \wd0{\hss\ensuremath{\mystrut^{\tt{\hspace{0.11cm}w%
  \hspace{-0.01cm}v}}}\hss}}\box0%
}}

\newcommand{\weakWVC}{{%
  \setbox0\hbox{\ensuremath{\leftlsquigarrow}}%
  \rlap{\hbox to \wd0{\hss\ensuremath{\mystrut^{\tt{\hspace{0.11cm}w%
  \hspace{-0.01cm}v}}}\hss}}\box0%
}}

\newcommand{\mycomment}[1]{%
  \textnormal{\small{\sf{#1}}}%
}
\newcommand{\mike}{{\hspace{-0.04cm}mike\hspace{-0.03cm}}}
\newcommand{\GPA}{{G\hspace{-0.04cm}P\hspace{-0.08cm}A}}
\newcommand{\tab}{\mbox{\hspace{0.25cm}}}

\title[Advances in ELP solvers]{\mbox{A survey of advances in epistemic logic
program solvers}}

\author[A. P. Leclerc{\small~and~}P. T. Kahl]{%
  ANTHONY P. LECLERC\\
  Space and Naval Warfare Systems Center Atlantic, North Charleston, SC, USA\\
  College of Charleston, Charleston, SC, USA\\
  \email{anthony.leclerc@navy.mil{\normalfont;~}leclerca@cofc.edu}\medskip\\
  {\normalfont\normalsize PATRICK THOR KAHL}\smallskip\\
  Space and Naval Warfare Systems Center Atlantic, North Charleston, SC, USA\\
  \email{patrick.kahl@navy.mil}
}

\submitted{9 April 2018}
\revised{19 April 2018}
\accepted{13 May 2018}

\begin{document}
\maketitle

%% ABSTRACT
\begin{abstract}
Recent research in extensions of Answer Set Programming has included a
renewed interest in the language of Epistemic Specifications, which adds modal
operators $\K$ (``known'') and $\M$ (``may be true'') to provide for more
powerful introspective reasoning and enhanced capability, particularly when
reasoning with incomplete information. An \emph{epistemic logic program} is a
set of rules in this language. Infused with the research has been the desire
for an efficient solver to enable the practical use of such programs for
problem solving. In this paper, we report on the current state of development
of epistemic logic program solvers.
\end{abstract}

%% KEYWORDS
\begin{keywords}
Epistemic Logic Program Solvers, Epistemic Specifications, Epistemic Logic
Programs, World Views, Solvers, Epistemic Negations, Answer Set Programming
Extensions, Logic Programming\vspace{-0.7cm}
\end{keywords}

%% INTRODUCTION
%----------------------------------%
% ELPSolverSurvey_Introduction.tex %
%----------------------------------%

\section{Introduction}\label{IntroductionSection}

In the study of knowledge representation and reasoning as related to logic
programming, the need for sufficient expressive power in order to correctly
represent incomplete information and perform introspective reasoning when
modeling an agent's knowledge of the world has been slowly realized. As such,
Michael Gelfond's language of \emph{Epistemic Specifications}
\cite{Gelfond91a,Gelfond94a} has seen renewed interest
\cite{FaberWoltran11a,Truszczynski11a,Gelfond11a,Kahl14a,%
KahlWatsonBalaiGelfondZhang15a,Su15a,FarinasHerzigSu15a,ShenEiter16a,%
ZhangZhang17a}. Much of the focus of late has been on semantic subtleties,
particularly for rules involving recursion through modal operators.
However, concomitant interest in the development of solvers for finding the
\emph{world views} (collections of \emph{belief sets} analogous to the answer
sets of an ASP program) of an epistemic logic program has progressed to the
point that a number of choices are now available: {\em ESmodels}
\cite{ESmodels}, {\em ELPS} \cite{ELPS}, {\em ELPsolve} \cite{ELPsolve},
{\em EP-ASP} \cite{EP-ASP}, {\em Wviews} \cite{Wviews}, {\em EHEX} \cite{EHEX},
and {\em selp} \cite{selp}. Additionally, {\em GISolver} \cite{GISolver} and
{\em PelpSolver} \cite{PelpSolver} are tools for finding the world views of
extensions of Epistemic Specifications that can also be used for epistemic
logic programs with minor syntactic translation. For awareness and to promote
continued research, development, and use of Epistemic Specifications and its
variants, we present a survey of epistemic logic program solvers.\footnote{%
DISCLAIMER:
The views and opinions expressed may not reflect those of the US Government.}

\newpage

The paper is organized as follows. In section 2 we provide a brief
overview of the language of Epistemic Specifications including a
synopsis of the syntax and semantics of the different versions
supported by the solvers included in this survey.  In section 3 we
discuss the solvers themselves and consider history,
influences, implementation, and key features. %possible motivation
%We also include comments to facilitate the usage of individual solvers based
%on our experiences.
In section 4 we include
performance data on extant solvers compiled from experiments on select
epistemic logic programs. We close with a summary and statements about
the future of ELP solvers.

%% EPISTEMIC SPECIFICATIONS
%---------------------------------------------%
% ELPSolverSurvey_EpistemicSpecifications.tex %
%---------------------------------------------%

\section{Epistemic Specifications}\label{EpistemicSpecificationsSection}

Gelfond presented the following example in \cite{Gelfond91a} to demonstrate
the need for extending the language of what we now call answer set programming
(ASP) in order to ``allow for the correct representation of incomplete
information in the presence of multiple answer sets.''

\smallskip
\indent \mycomment{\% rules for scholarship eligibility at a certain college}\\
\indent $eligible(S) \leftarrow high\GPA(S).$\\
\indent $eligible(S) \leftarrow fair\GPA(S),~minority(S).$\\
\indent $\neg eligible(S) \leftarrow \neg high\GPA(S),~\neg fair\GPA(S).$\\
\indent \mycomment{\% ASP attempt to express an interview requirement when
eligibility cannot be determined}\\
\indent $interview(S) \leftarrow \ASPnot eligible(S),
                                 \ASPnot \neg eligible(S).$\\
\indent \mycomment{\% applicant data}\\
\indent $fair\GPA(\mike) \ASPor high\GPA(\mike).$
\smallskip

\noindent
This program correctly computes that the eligibility of Mike is indeterminate,
but its answer sets, $\{fair\GPA(\mike),interview(\mike)\}$ and
$\{high\GPA(\mike),eligible(\mike)\}$, do not conclude that an interview is
required since only one contains $interview(\mike)$.

Gelfond's solution was to extend ASP by adding modal operator $\K$ (``known'')
and changing the fourth rule above to:
\smallskip

\indent \mycomment{\% updated rule to express interview requirement using modal
operator $\K$}\\
\indent $interview(S) \leftarrow \ASPnot\K\ eligible(S),
                                 \ASPnot\K\ \neg eligible(S).$

\smallskip
\noindent
The updated rule means that $interview(S)$ is true if both
$eligible(S)$ and $\neg eligible(S)$ are each \emph{not known} (i.e., not in
all belief sets of the world view).

The new program has a world view with two belief sets:
$\{fair\GPA(\mike),interview(\mike)\}$ and
$\{high\GPA(\mike),eligible(\mike),interview(\mike)\}$,
both containing $interview(\mike)$. It therefore correctly entails that Mike
is to be interviewed.

%Those familiar with the ASP notion of \emph{cautious entailment}\footnote{Given
%ASP program $\mathcal{P}$, literals in the intersection of all answer sets of
%$\mathcal{P}$ are considered \emph{cautiously entailed} by $\mathcal{P}$.} may
%see the $\K$ modal operator as a function of cautious entailment. 

Since its 1991 introduction, four revisions of the language of Epistemic
Specifications have been implemented in solvers.  Other revisions of Epistemic
Specifications have been proposed \cite{FarinasHerzigSu15a,ZhangZhang17a}, but
to the best of our knowledge, no solvers for those versions were implemented.
The revision we call \textbf{\emph{ES1994}} is described in \cite{Gelfond94a,%
BaralGelfond94a}. With a renewed interest in Epistemic Specifications nearly
two decades later, Gelfond proposed an update \cite{Gelfond11a} to the language
in an attempt to avoid unintended world views due to recursion through modal
operator $\K$.  We refer to this version as \textbf{\emph{ES2011}}. Continuing
with Gelfond's efforts to avoid unintended world views due to recursion, but
through modal operator $\M$, Kahl proposed a further update \cite{Kahl14a}. We
refer to this version as \textbf{\emph{ES2014}}. Most recently, Shen and Eiter
proposed yet another update \cite{ShenEiter16a} to address perceived issues
with unintended world views remaining in the language.  We call this version
\textbf{\emph{ES2016}}.

A synopsis of the syntax and semantics of the different versions of
Epistemic Specifications covered by the surveyed solvers is given below.
We encourage the reader to see the papers previously referenced for more
detailed discussions of individual language versions.

In general, the syntax and semantics of Epistemic Specifications follow those
of ASP with the notable addition of modal operators $\K$ and $\M$ and the new
notion of a \emph{world view}. A world view of an ELP is a collection of
\emph{belief sets} (analogous to the answer sets of an ASP program) that
satisfies the the rules of the ELP and meets certain other requirements as
given in the table below.

\noindent\hrulefill

\noindent
{\footnotesize
\textbf{Syntax}\\
An epistemic logic program (ELP) is a set of rules in the language of
Epistemic Specifications, a rule having the form
$$\ell_1\ASPor ... \ASPor\ell_k\leftarrow e_1,~...,~e_n.$$
\noindent
where $k\ge 0$, $n\ge 0$, each $\ell_i$ is a \emph{literal} (an atom or a
classically-negated atom; called an \emph{objective literal} when needed to
avoid ambiguity), and each $e_i$ is a literal or a \emph{subjective literal} (a
literal immediately preceded by $\K$ or $\M$) possibly preceded by
$\leftASPnot$(default negation).\footnote{In ES1994 and ES2011, negated
subjective literals have their modal operators prefaced with ~$\neg$~ rather
than ~{\tt not}. In the semantics given above, we extend the syntax by allowing
default-negated literals to follow modal operator ~$\K$~ and consider
~$\M\hspace{0.05cm}\ell$~ to be simply a shorthand for
~$\leftASPnot\K\hspace{0.05cm}\leftASPnot\ell$~
(or ~$\neg\K\hspace{0.05cm}\leftASPnot\ell$~ in ES1994/ES2011 syntax).}
As in ASP, a rule having an objective/subjective literal with
a variable term is a shorthand for all ground instantiations of the rule. The
~$\leftarrow$~ symbol is optional if the body of the rule is empty (i.e.,
$n{=}0$).

\noindent
\textbf{When a Subjective Literal Is Satisfied}\vspace{-0.45cm}
}
\begin{figure}[h!] %[!t]
{\centering
\includegraphics[height=9cm,width=13.475cm]%
{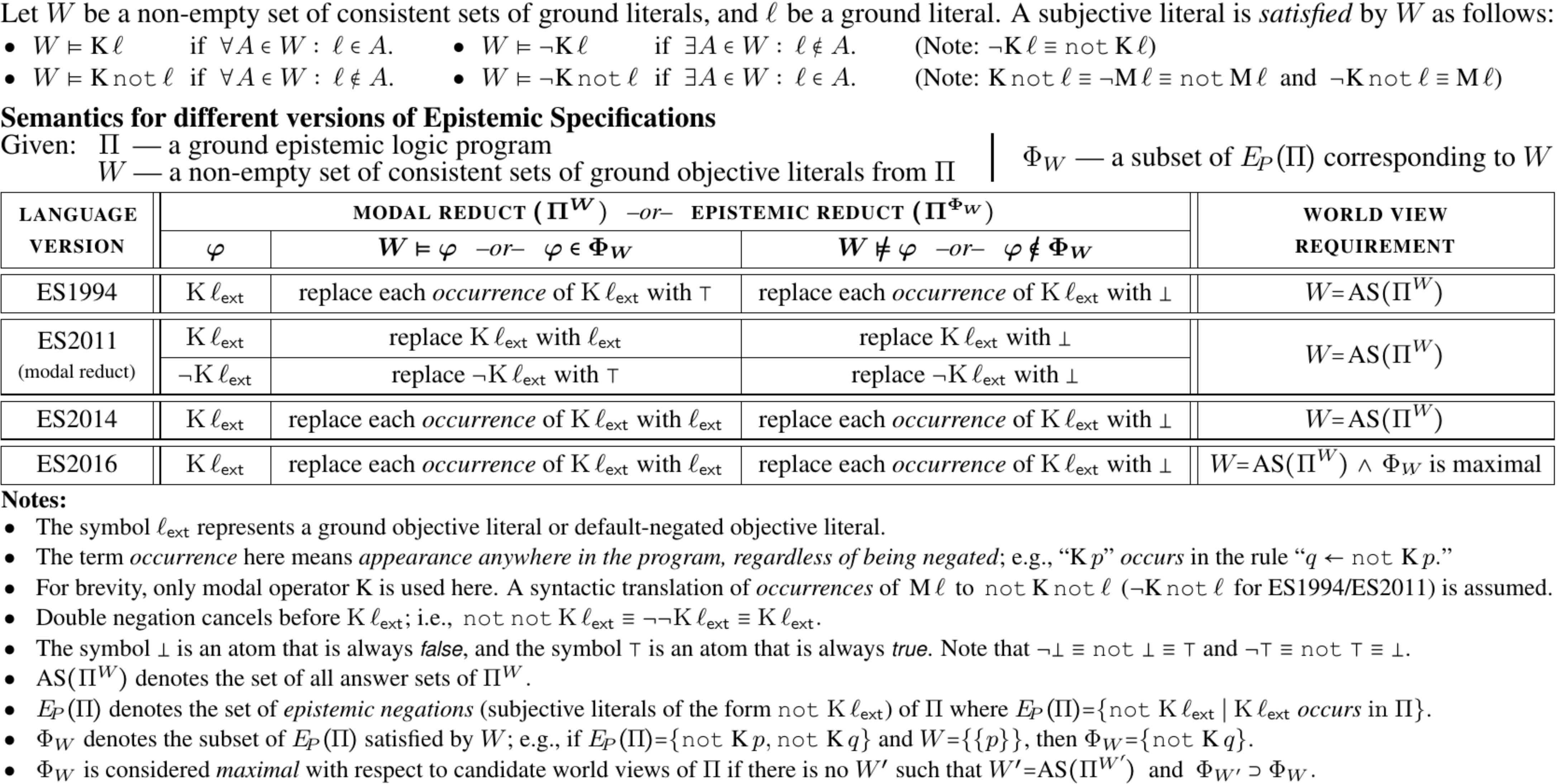}
}
\end{figure}

\vspace{-1.25cm}
\noindent\hrulefill
\vspace{-0.25cm}

%% SOLVERS
%-----------------------------%
% ELPSolverSurvey_Solvers.tex %
%-----------------------------%

\section{Solvers}\label{SolversSection}

In the subsections below we discuss the ELP solver development efforts
spanning, in chronological order, the years from 1994 to 2018.
Included in the group are two solvers, {\em GISolver} and
{\em PelpSolver}, which were designed for different extensions of
ASP, but nevertheless are able to compute the world views of ELPs given
simple translations of the input language encoding.

We note that all of the extant solvers discussed operate from the
command line, which is to say that no Integrated Development
Environment (IDE) or Graphical User Interface (GUI) currently exists
for solving ELPs.  We also note that all extant ELP solvers generate
what can be called an \emph{epistemic reduct framework} for the ELP.
This is a core ASP program that when instantiated with a ``guess''
(truth value assignments for the subjective literals represented by a
subset of the epistemic negations that are considered \true) will
correspond to the \emph{epistemic reduct} for that guess. An underlying
(or background) ASP solver such as {\em DLV} \cite{DLV},
{\em DLVHEX2} \cite{DLVHEX2}, {\em claspD}, or
{\em clingo} \cite{Potassco}) is then used to compute the answer sets
of the epistemic reduct.

The terms ``loosely coupled'' and ``tightly coupled'' are used in our
discussions of the implementations of the solvers.  By loosely coupled
we mean that the underlying ASP solver is invoked as a separate
process rather than through a library with a specific Application
Programming Interface (API).  A loosely coupled implementation has the
advantage that it can be easily modified to utilize a different
underlying ASP solver, assuming the capabilities and input language
syntax of the ASP solvers are similar.  A tightly coupled
implementation is not as flexible but generally more efficient,
as it avoids the overhead of creating and communicating with a
separate process.

The input language of a given solver is typically a subset of the ASP Core 2
standard \cite{ASPCore2Standard} with the addition of modal operators $\K$ and
$\M$. For example, the ``$\leftarrow$'' symbol is typically represented by the
2-character string ``{\tt :-}'' though some solvers may accept other
representations. {\em ELPsolve} and {\em EP-ASP} rely on {\em ELPS} for
preprocessing the input program, which requires additional statements in the
program to explicitly define the domain for predicate terms as a \emph{sorted
signature}. The input language of {\em ELPS} also uses ``{\tt K\$}'' and
``{\tt M\$}'' to represent modal operator symbols ``$\K$'' and ``$\M$''
(respectively). The {\em selp} system accepts the same input language as
{\em ELPS}, but does not depend on {\em ELPS} for processing. It can
alternatively accept ``{\tt \$not\$}'' as the \emph{epistemic negation
operator}, which is equivalent to ``$\ssleftASPnot\K$'' in our notation.
We refer the reader to documentation and example programs available with the
solver distributions for specifics on the individual input languages. We will
continue to use the notation described in
Section~\ref{EpistemicSpecificationsSection} with the understanding that it
differs from the actual input languages of the various solvers.

\begin{figure}[h!]
{\centering
\caption{ELP Solver Contact and Download Information}\label{solverContacts}
\includegraphics[height=5cm,width=13.5cm]%
{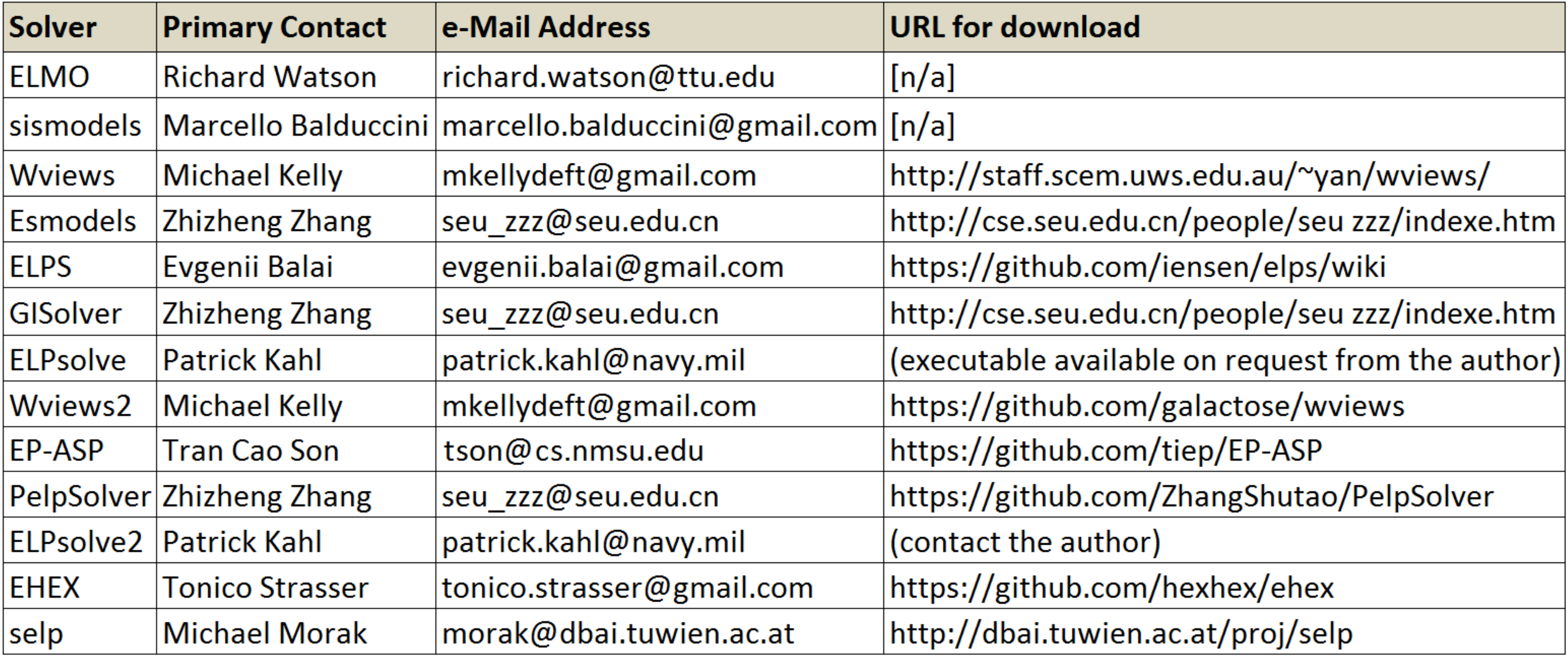}
}
\end{figure}

Near the end of the paper are a number of summary tables. These include a
historical synopsis of solver development (Table~\ref{solverSummary}), a
brief summary of solver features (Table~\ref{solverFeatures}), and a listing
of solver contacts \& download information (Table~\ref{solverContacts}).

\vspace{-0.2cm}
\subsection{ELMO}

The earliest work on the development of an ELP solver was that of Richard
Watson in 1994 while a graduate student of Michael Gelfond when he was at
the University of Texas at El Paso.
Though not a solver per se, Watson's {\em ELectronic MOnk} ({\em ELMO}) was
a Prolog implementation of an inference engine for a limited class of ELPs.
{\em ELMO} also required the {\em SLG} system developed at Southern Methodist
University and State University of New York (SUNY) at Stony Brook
\cite{ChenWarren93a}.
There is no extant electronic binary or source; however, the printed source code
is listed as an appendix of Watson's master's thesis.

In his thesis, Watson demonstrates the efficacy of {\em ELMO} by reporting the
answers to queries using ELMO for various examples, including the scholarship
eligibility problem of Section~\ref{EpistemicSpecificationsSection}.

\vspace{-0.2cm}
\subsection{sismodels}

In 2001, Marcello Balduccini, working as a graduate student with Michael
Gelfond at Texas Tech University, began work on a solver that extended
{\em Smodels} \cite{Simons00a,Smodels} with strong introspection.
He called his solver {\em sismodels}.  The work, however, never progressed
beyond proof-of-concept.  As with {\em ELMO}, there is no extant electronic
binary or source for {\em sismodels}. It is included here as it is the first
known attempt to implement an ELP solver in the sense that its output was the
world views of the input ELP.

\vspace{-0.2cm}
\subsection{Wviews}

Working with Yan Zhang as his advisor for his honours thesis \cite{Kelly07a} at
the University of Western Sydney, Michael Kelly implemented an ES1994 solver
{\em Wviews} based on the algorithm suggested in \cite{Zhang06a}. Kelly's
implementation features a grounder and a solver in a single executable that is
loosely coupled with {\em DLV} as the background ASP solver. This
was the first general epistemic logic program solver, and it is still
available as a Microsoft Windows executable. Although the original C++ source
code for this version of the solver was lost, Kelly has recently posted a
Python version of {\em Wviews} \cite{Wviews} that we will refer to as
{\em Wviews2}. This new version contains ``major modifications'' according
to its author.

{\em User Experience:} {\em Wviews2} is the one to use for ES1994 semantics.
We note that {\em Wviews2} tries one guess at a time, which can result
in calling the underlying ASP solver $2^k$ times, where $k$ is the
number of epistemic negations, limiting its practical use to relatively
small (w.r.t. the number of epistemic negations) ELPs. Overcoming this
limitation is a challenge for all solver developers. {\em Wviews2}
exhausts the search space iteratively to ensure all world views are
computed.

\vspace{-0.2cm}
\subsection{ESmodels}

After spending the summer of 2011 at Texas Tech University, Zhizheng Zhang
returned to Southeast University with the idea of implementing a solver for
Gelfond's new version of Epistemic Specifications, ES2011. He started with
a grounder, and by 2012 had implemented (with the help of graduate students
Rongcun Cui and Kaikai Zhao) {\em ESParser} \cite{CuiZhangZhao12a}. This was
followed by {\em ESsolve} in 2013, resulting in a grounder-solver system they
called {\em ESmodels} \cite{ZhangZhaoCui13a}. {\em ESsolve} is loosely
coupled with ASP solver {\em claspD}.

Although work on {\em ESmodels} continued for a short time \cite{ZhangZhao14a},
the system is available today only as a Microsoft Windows executable from
Zhang's homepage at Southeast University. It is the only ES2011 solver known.

{\em User Experience:} {\em ESmodels} appears to work reasonably well with
programs that are relatively small w.r.t. the number of epistemic
negations. With larger programs, we sometimes observed a runtime error
or the unexpected result of no world views for programs known to be
consistent.

We note that the $\M$ modal operator is not directly
supported; however, equivalent\footnote{Equivalence here is with respect
to the world views of respective programs, modulo any fresh atoms
introduced.} constructs can be created by
replacing each \emph{occurrence} of $\Mell$ as follows:\\
\tab ~1.~ Replace $\Mell$ with $\neg\Kell^\prime$ where $\ell^\prime$ is a
fresh atom. (Remove any double negation before $\K$.)\\
\tab ~2.~ Add the following new rule: ~$\ell^\prime\leftarrow\ssASPnot\ell.$\\
\noindent
Classical/strong negation is also not directly supported other
than to denote a negated subjective literal, but, as before, a workaround
exists by replacing each occurrence of $\neg\ell$ as follows:\\
\tab ~1.~ Replace $\neg\ell$ with $\ell^\prime$ where $\ell^\prime$ is a
fresh atom.\\
\tab ~2.~ Add the following constraint: ~$\leftarrow\ell,~\ell^\prime.$

\vspace{-0.2cm}
\subsection{ELPS}

As graduate students at Texas Tech University, Evgenii Balai and
Patrick Kahl worked together on a version of Epistemic Specifications
that uses a sorted signature. A program written in this version is
called an \emph{epistemic logic program with sorts} \cite{BalaiKahl14a}.
This effort was strongly influenced by Balai's work on {\em SPARC}
\cite{BalaiGelfondZhang13a}, a version of the language of ASP using a
sorted signature. Balai implemented the ES2014 (with sorted signature)
solver {\em ELPS} using an algorithm formed by
combining Kahl's ES2014 algorithm with Balai's SPARC algorithm. Much
of the Java code from an old version of SPARC was able to be reused,
allowing Balai to create a working solver in about three days worth of
work---an impressive feat. {\em ELPS} is loosely coupled with the ASP
solver {\em clingo}.

Although {\em ELPS} is a stable, reliable ES2014 solver for small
(in number of epistemic negations) programs that makes only one call
to the underlying ASP solver, its memory requirements can grow
exponentially with the number of epistemic negations
\cite{KahlLeclercSon16a}. It does, however, provide a nice front end
for other solvers, such as {\em  ELPsolve} and {\em EP-ASP}, to be able
to translate an ELP with sorts into an ASP epistemic reduct framework.
Java source code and a pre-built {\em .jar} file are available.

{\em User Experience:} {\em ELPS} works very well for programs that
are relatively small with respect to the number of epistemic negations,
but due to exponentially-growing memory needs as the number of
epistemic negations grow, it has limited application as a solver.
Nonetheless, it is one of the only solvers with a detailed user manual.
We note that it outputs all world views of its input program with no
option for changing this.
It does have the option ``-o'' for outputting a file representing the
epistemic reduct framework of the input program, along with rules for
generating all combinations of subjective literal truth values. This gives
{\em ELPS} potential value as a front end for other solvers. 

\vspace{-0.2cm}
\subsection{GISolver}

Zhizheng Zhang and graduate students Bin Wang and Shutao Zhang embarked on
developing the solver {\em GISolver} for an extension of ASP called
{\em GI-log} \cite{ZhangWangZhang15a}. {\em GISolver} can be used to find
world views of ES2014 programs after minor syntactic translations. It is
loosely coupled with {\em clingo} as the underlying ASP solver.
Like {\em ESmodels}, this solver is currently available only as a Microsoft
Windows executable from Zhang's homepage at Southeast University. It appears
to have been a stepping stone in the development of {\em PelpSolver}
discussed later.

{\em User Experience:} {\em GISolver} works well for relatively small (w.r.t.
the number epistemic negations) ELPs provided they are appropriately translated
to GI-log syntax by converting subjective literals as shown below:\smallskip\\
\indent
\begin{oldtabular}{lcl}
{\bf ES2014 syntax} &               & {\bf GI-log syntax}\vspace{-0.2cm}\\\hline
{\tt ~~~~ K p}      & $\Rightarrow$ & {\tt ~K[1,1] p}\\
{\tt ~not K p}      & $\Rightarrow$ & {\tt ~K[0,1) p}\\
{\tt ~~~~ M p}      & $\Rightarrow$ & {\tt ~K(0,1] p}\\
{\tt ~not M p}      & $\Rightarrow$ & {\tt ~K[0,0] p}\\
\end{oldtabular}\smallskip

\vspace{-0.2cm}
\subsection{ELPsolve}

{\em ELPsolve} was developed in 2016 by the authors. Two primary efficiency
goals were pursued: (1) develop an ELP solver that avoids the large memory
requirements of {\em ELPS}; and (2) parallelize the solver to take advantage
of multi-core processors. Other goals included
support for the updated semantics of Shen \& Eiter (ES2016) and
optimization for conformant planning.  To solve the memory issue, {\em
ELPsolve} partitions guesses into fixed-sized groups, rather than computing
all guesses with one ASP solver call. These groups are systematically
generated in an order that guarantees the maximality requirement of
ES2016 and permits pruning of the search space when multiple world
views are desired. Groups of guesses are mutually exclusive so that
parallelization can occur with minimal synchronization.
{\em ELPsolve} supports both ES2014 and ES2016 semantics. Binary
executables for Windows, Mac, and Linux are available upon request.

{\em User Experience:} {\em ELPsolve} has several options, including the
ability to specify the (maximum) number of world views to output, the number
of processors to be used, conformant planning mode (with planning horizon),
and a configuration file.  The configuration file is used to specify less
volatile configuration options such as group size, language semantics to use
(ES2014 or ES2016), and ASP solver path. {\em ELPsolve} itself is invoked
from a script which first seamlessly calls ELPS for translating the ELP
(with sorts) input program into an epistemic reduct framework, then
invokes ASP grounder {\em gringo} to ground the program, and finally calls
{\em ELPsolve} for further processing. {\em ELPsolve} is loosely-coupled
with {\em clingo} for backend ASP program solving.

\vspace{-0.2cm}
\subsection{EP-ASP}

Tran Cao Son worked as an Office of Naval Research faculty researcher at
Space and Naval Warfare Systems Center Atlantic in the summer of 2016.
His work with the authors on the development of {\em ELPsolve} stimulated
his interest and led to his own approach, resulting in a new solver:
{\em EP-ASP}. The core idea of this solver is to take the epistemic reduct
framework (as in {\em ELPS} and {\em ELPsolve}), but instead of solving
for all possible guesses at once (like {\em ELPS}) or systematically in
groups of guesses (like {\em ELPsolve}), it uses the underlying ASP solver
to compute a single answer set. Due to the way the epistemic reduct
framework is constructed, this answer set represents a \emph{consistent}
guess (i.e., one that results in a consistent epistemic reduct). The framework
is instantiated for that guess, all answer sets are computed, and the answer
sets are checked to see if they represent a world view. A constraint is then
added to eliminate this guess from further consideration, and the process is
repeated until all world views of the program are discovered.

For input to {\em EP-ASP}, an epistemic reduct framework representation of the
ELP is created first using {\em ELPS}. {\em EP-ASP} works completely within the
{\em clingo} runtime environment, using embedded Python to control iteration in
a \emph{multi-shot ASP solving} approach
\cite{GebserKaminskiKaufmannSchaub17a}.
After creating a proof-of-concept version for ES2014, Son enlisted the aid of
his New Mexico State University graduate student Tiep Le to implement
support for ES2016, the use of brave and cautious reasoning for pruning
the search space, and optimizations for conformant planning.

The solver supports both ES2014 and ES2016 semantics and is among the fastest
solvers for the sample programs used in our tests.

{\em User Experience:} {\em EP-ASP} has several options, including the
ability to specify the use of brave and cautious consequences as a preliminary
step to prune the search space, language semantics to use (ES2014 or ES2016),
and conformant planning mode.

\vspace{-0.2cm}
\subsection{PelpSolver}

Continuing with the success of {\em GISolver}, Zhizheng Zhang and
Shutao Zhang developed a solver for probabilistic-epistemic logic programs
\cite{ZhangZhang17a} called {\em PelpSolver}.
With appropriate syntactic translation, {\em PelpSolver} can be used to solve
ES2016 programs. It is implemented in Java and is loosely coupled with
{\em clingo} as the underlying ASP solver.

The development of the language of probabilistic-epistemic logic programs was
a culmination of language extensions that were positively influenced by ELP
solver development. During development of {\em ESmodels}, implementation of
the world view verification step involved counting the number of occurrences,
$count(\ell)$, of the objective literal part, $\ell$, of each subjective
literal in the computed belief sets. For example, if checking subjective
literals against a set of, say, $5$ belief sets, to verify
$\K\hspace{0.05cm}p$,
\hspace{0.03cm}$count(p)\hspace{-0.04cm}=\hspace{-0.04cm}5$ is required, to
verify $\M\hspace{0.05cm}q$,
\hspace{0.03cm}$count(q)\hspace{-0.04cm}\ge\hspace{-0.04cm}1$ is required, and
so forth. They observed that other numbers/number ranges could easily be
checked, leading to the realization that the ability to specify the
\emph{fraction} of belief sets required to contain a particular literal might
be useful for modeling certain problems. This led to the new language
extensions.

{\em User Experience:} {\em PelpSolver} comes with a pre-built
{\em .jar} file, but can also be built using a Maven {\em pom.xml} file.
One command-line option exists for optimization.  The conversion from
an ELP program to a probabilistic-epistemic logic program is the same as
that given for {\em GISolver}.

\vspace{-0.2cm}
\subsection{ELPsolve2}

{\em ELPsolve2} was developed in 2017 by the authors.  Unlike {\em
 ELPsolve}, this version of the software has not been officially
released to the public, nor have there been any technical papers
written about it.  For this reason we describe {\em ELPsolve2} in a
little more detail for this survey.

Two primary design goals guided the development of {\em ELPsolve2}:
efficiency and support for additional features.  Specifically, {\em ELPsolve2}
improves on {\em ELPsolve} in five ways:
\begin{itemize}\vspace{-0.15cm}
\item replaces ``loosely coupled'' ASP solver interaction with
  ``tightly coupled'' interaction
\item implements an ``invalid guess'' filter
\item uses brave and cautious reasoning to reduce the number of
  epistemic negations
\item improves the optimization used for conformant planning problems
\item implements {\em World View Constraints} (WVCs)
\end{itemize}\vspace{-0.15cm}

Both {\em ELPsolve} and {\em ELPsolve2} utilize the {\em clingo}
ASP solver for solving the epistemic reduct framework.  With {\em
 ELPsolve}, calls to {\em clingo} are performed as external processes
that require time to instantiate.  Furthermore, these processes
communicate results less efficiently through the operating system.
Instead, {\em ELPsolve2} utilizes {\em clingo's} {\em C}
programming language interface.  Time to invoke a {\em clingo} call
and store the results is therefore reduced.

We call a guess that contains epistemic negations that cannot co-exist an
``invalid guess.''  {\em ELPsolve2} filters such guesses, thus avoiding
unnecessary computation. The following pairs of epistemic negations cannot
co-exist:
\begin{itemize}\vspace{-0.15cm}
\item $\K\ \ell$ ~and~ $\leftASPnot\M\hspace{0.085cm}\ell$
\item $\K\ \ell$ ~and~ $\M\hspace{0.085cm}\overline{\ell}$
\item $\K\ \ell$ ~and~ $\K\ \overline{\ell}$
\end{itemize}\vspace{-0.15cm}
\noindent
where ~$\overline{\ell}$~ denotes the logical complement of $\ell$.
For example, if ~$\ell=\neg p$~ then $\overline{\ell}=p$.

Brave and cautious reasoning was first successfully used in {\em EP-ASP} to
reduce the number of epistemic negations under consideration, pruning the
search space for certain ELPs.
{\em ELPSolve2} incorporates this optimization.  We note that for some
problems, brave and cautious reasoning yields no reduction (e.g.,
conformant planning problem); however, for others a considerable
reduction is achieved (e.g., scholarship eligibility problem).

{\em ELPsolve2} improves the optimization for conformant planning problems
over {\em ELPsolve} by further reducing the search space based on the
assumption that only one action is performed at each step. Although this
assumption may seem too constraining, optimizations related to conformant
planning are highly specialized and can result in dramatic improvements in
performance when applied as intended.

Finally, {\em ELPsolve2} allows for the extension known as \emph{world view
constraints} proposed by the authors in \cite{KahlLeclerc18a}. This has the
potential for reduction of the search space over encodings that do not use
world view constraints. Thus, from a solver perspective, this can be viewed
as a general approach with the potential for performance improvement rather
than an optimization applicable only to very specific applications such as
conformant planning.

{\em User Experience:} {\em ELPsolve2} comes with all the options from
{\em ELPsolve}, and adds options for brave and cautious reasoning as
well as different output formats.

\vspace{-0.2cm}
\subsection{EHEX}

At the time of this writing, Anton ``Tonico'' Strasser is a graduate
student at TU Wien working under the advisement of Thomas Eiter and
Christoph Redl.  His ES2016 solver {\em EHEX} adds epistemic negations
to HEX programs, which allows integration of external computation
sources. {\em EHEX} works with {\em DLVHEX2} as the underlying ASP
solver, but uses {\em clingo} as well to perform optional brave and
cautious reasoning.  {\em EHEX} is written in Python and is loosely
coupled with the ASP solver.

% EHEX utilizes clingo as well, at least in order to perform optional brave
% and cautious reasoning.  EHEX is loosely coupled.  They apply our rules:
% see   https://github.com/hexhex/ehex/blob/master/ehex/translation.py
% They appear to have a conformant planning module.

{\em User Experience:} {\em EHEX} has a number of options and many example
programs are available on the developer's GitHub page. Given an already
existing installation of {\em DLVHEX2} with the {\em NestedHexPlugin},
{\em EHEX} builds and installs easily. However, we found it challenging to
build {\em DLVHEX2} with the {\em NestedHexPlugin} from source. Even though it
is a work-in-progress as of this writing, {\em EHEX} performed quite well. We
look forward to further developments.

\vspace{-0.2cm}
\subsection{selp}

Another graduate student at TU Wien, Manuel Bichler, working under the
advisement of Stefan Woltran and Michael Morak, applied ASP rule
decomposition \cite{BichlerMorakWoltran16a} to ELP solving to develop a
single-shot (w.r.t. ASP solver calls) epistemic logic program solver called
{\em selp}.
%The solver implements the EFLP semantics described in
%\cite{ShenEiter16a} which can be viewed as ES2016 using FLP semantics
%\cite{FaberPfeiferLeone11a} for nested default negation
%(``{\tt not\hspace{0.07cm}not}\hspace{-0.02cm}'').\footnote{Nested default
%negation cancels out in FLP semantics. For example, the one-rule ASP program
%$\{p\leftarrow\ssleftASPnot\leftASPnot p\}$ has one answer set $\{\}$ under
%FLP semantics as it is equivalent to $\{p\leftarrow p\}$. The same program has
%the additional answer set $\{p\}$ per \cite{LifschitzTangTurner99a} as it is
%equivalent to$\{p\hspace{-0.02cm}\leftarrow\hspace{-0.02cm}\leftASPnot
%p^\prime\hspace{-0.02cm},
%~p^\prime\hspace{-0.02cm}\leftarrow\hspace{-0.02cm}\leftASPnot p\}$
%with answer sets modulo $p^\prime$.}
The {\em selp} system is loosely coupled with {\em clingo}, and uses the
{\em lpopt} tool \cite{Bichler15a} to efficiently decompose ``large''
logic programming rules into smaller rules with the expectation
that such rules are more manageable/easier for {\em clingo} to handle.

{\em User Experience:} The {\em selp} system includes a number of Python
scripts, including its own tool for processing an input epistemic logic
programs with sorts. It generates rules containing a relatively large number
of body literals. The intent is to optimize the rules for decomposition
using the {\em lpopt} tool. This approach appears to work quite well for
certain programs (e.g., the scholarship eligibility problem described in
Section~\ref{EpistemicSpecificationsSection}) based on our experiments. It
also appears to benefit from the use of multiple threads with the backend
ASP solver {\em clingo}.

\vspace{-0.2cm}
\subsection{Solver Summary}

\begin{figure}[h!]
{\centering
\caption{Epistemic Logic Program Solvers}\label{solverSummary}
\includegraphics[height=6.5cm,width=13.5cm]%
{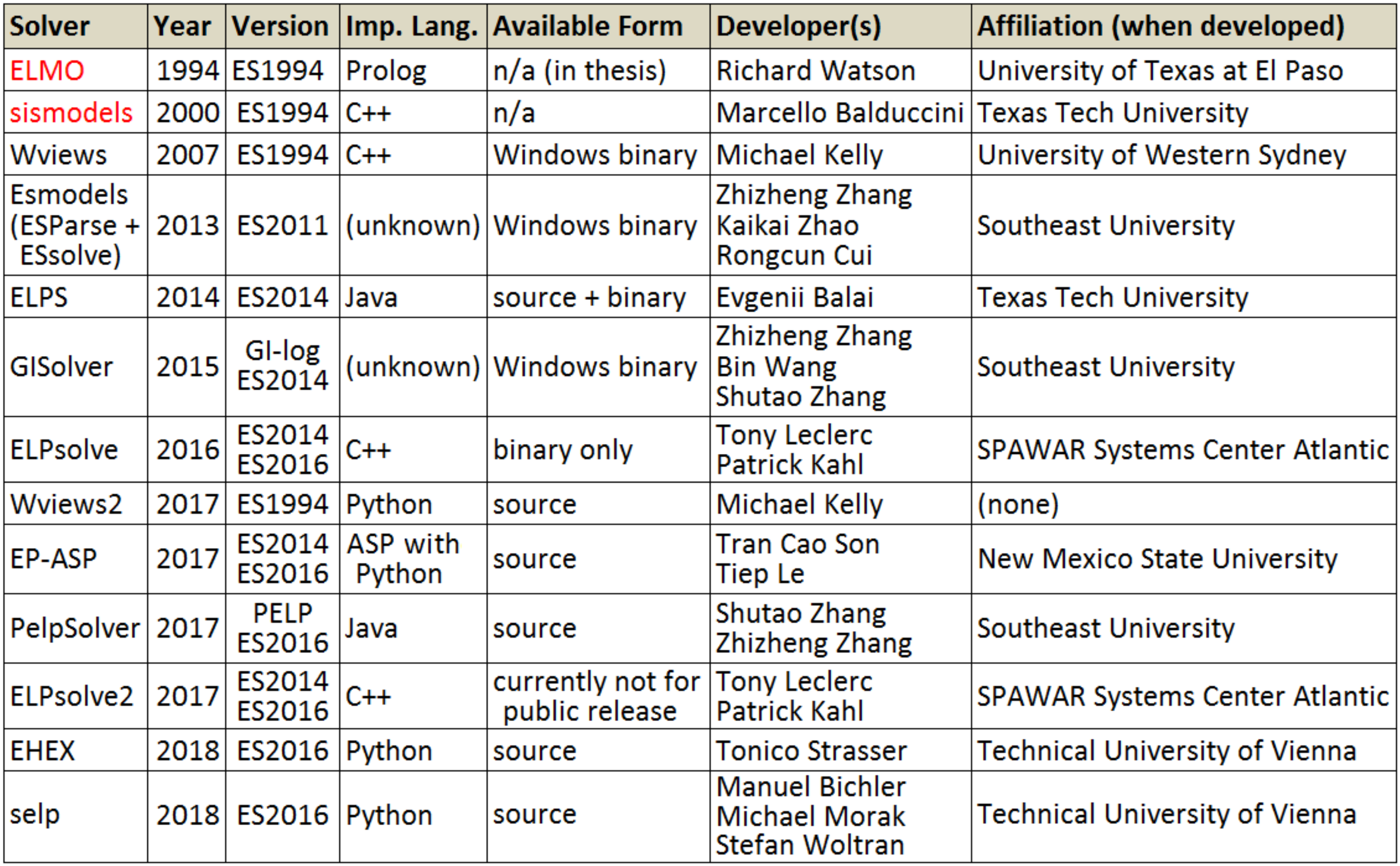}
}
\end{figure}

\begin{figure}[h!]
{\centering
\caption{ELP Solver Features}\label{solverFeatures}
\includegraphics[height=6.5cm,width=13.48cm]%
{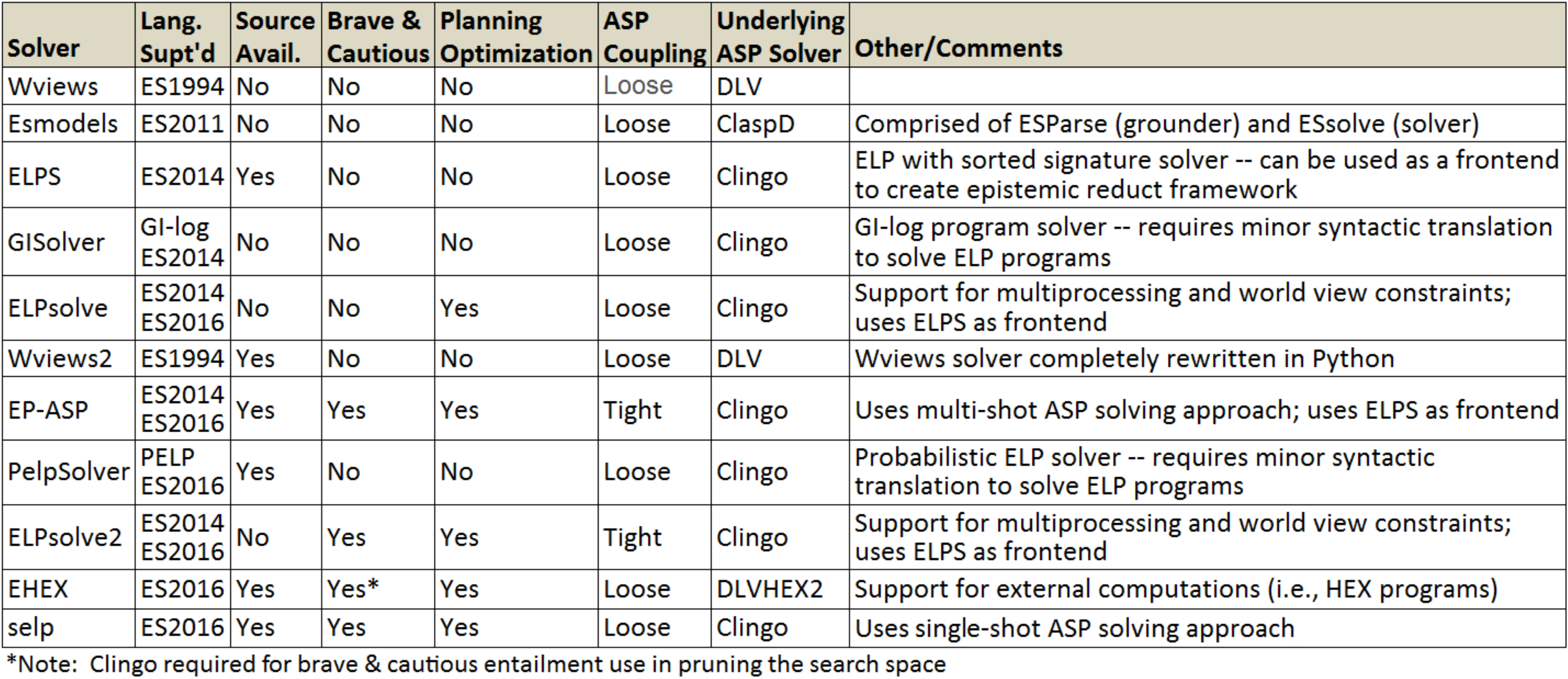}
}\vspace{-0.5cm}
\end{figure}

Table \ref{solverSummary} provides a general summary of all known ELP solvers.
{\em ELMO} and {\em sismodels} are highlighted in red to
indicate they no longer exist. 
Table \ref{solverFeatures} shows some of the key features of the ELP solvers
included in the performance experiments discussed in the next section.

%% EXPERIMENTS
%---------------------------------%
% ELPSolverSurvey_Experiments.tex %
%---------------------------------%

\section{Experiments}\label{ExperimentsSection}

Three epistemic logic programs of various sizes (w.r.t. the number of epistemic
negations) were used to test the capabilities and performance of different
solvers.
The {\em elig{\tt NN}} programs are instances of the scholarship eligibility
example described in Section~\ref{EpistemicSpecificationsSection}, where
{\tt NN} indicates the number of applicants.
The {\em yale{\tt N}} programs are instances of a variation of the Yale
shooting problem \cite{HanksMcDermott87a} encoded as describe in
\cite{KahlWatsonBalaiGelfondZhang15a}, where {\tt N} indicates the plan
horizon.
The {\em art\hspace{0.01cm}{\tt N}} programs are instances of a scalable
artificial problem we constructed involving combinations of both K and M modal
operators, where {\tt N} is the scaling factor.
Program listings are not included due to space constraints but are available
upon request.

\begin{figure}[h!]
{\centering
\caption{Experimental Results (total elapsed time in seconds for best
run)}\label{solverResultsTable}
\includegraphics[height=5.25cm,width=13.5cm]%
{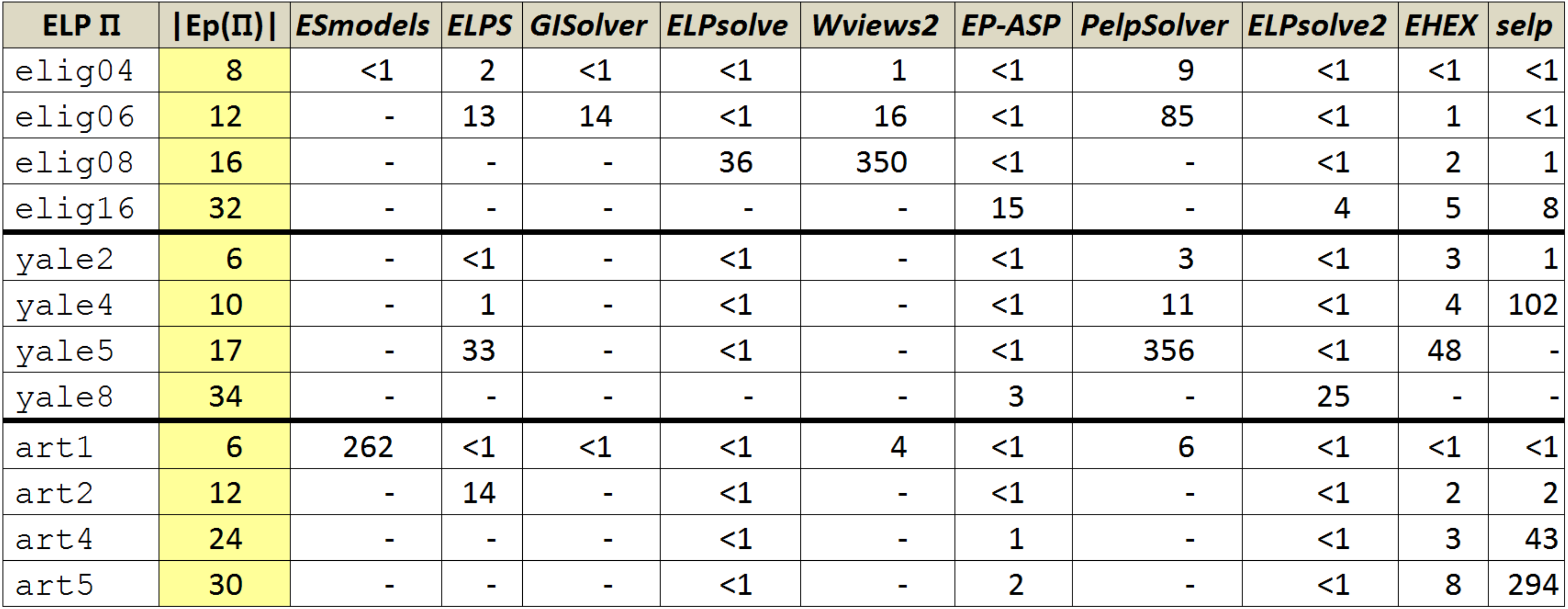}
}
\end{figure}

The test machine has an Intel i7 820QM @ 1.73 GHz processor with 8 GB
RAM. {\em ESmodels} and {\em GISolver} were run using a 64-bit Windows 10
operating system. All other solvers were run using a 64-bit Ubuntu 16.04
(Linux) operating system. {\em ELPsolve} and {\em EP-ASP} use {\em ELPS} to
create an epistemic reduct framework file from the input ELP (with sorts) file.
Table~\ref{solverResultsTable} shows the runtime results (in seconds) for
our tests. Times reported are for the entire solving experience, including (as
appropriate) time for creating the epistemic reduct framework file, time for
grounding, and time for displaying the results to the screen. Shell scripts
were used as warranted to minimize delay between processing steps. A dash ('-')
indicates that the solver was unable to solve the ELP on our system within 10
minutes (600 seconds).

The results indicate that the use of brave and cautious entailment by
{\em ELPsolve2}, {\em EP-ASP}, and {\em EHEX} have the potential to improve
performance dramatically for input similar to the {\em elig{\tt NN}} programs.
The approach used by {\em selp} also appears quite effective for programs of
this type.
For the {\em yale{\tt N}} programs, results are skewed in favor of solvers
with special optimizations for conformant planning problem encodings. It is
also apparent that solvers supporting ES2016 have an advantage for the
{\em art\hspace{0.01cm}{\tt N}} programs as solutions are found early, i.e.,
when all or most of the epistemic negations are true. Although we included
{\em GISolver} and {\em PelpSolver} in our tests, we note that these solvers
were designed for languages where Epistemic Specifications is but a subset.

%% CONCLUSIONS AND FUTURE WORK
%-------------------------------------------%
% ELPSolverSurvey_ConclusionsFutureWork.tex %
%-------------------------------------------%

\section{Conclusions}\label{ConclusionsFutureWorkSection}

Work on epistemic logic program solvers is clearly active.  We have
reviewed a number of solvers, most of which were developed within the
last five years. Significant improvements in both performance and the
ability to solve harder (w.r.t. the number of epistemic negations)
programs are evident.  The development of efficient and easier to use
solvers have allowed experimentation with different problems, syntax,
and semantics, and have in fact been useful to reveal and assess
different consequences of language variants.

Other ideas for improving performance include the use of world view
constraints, which have the potential to reduce the number of epistemic
negations \cite{KahlLeclerc18a}. For many solvers the search space of
epistemic negations can be partitioned into mutually exclusive
(independent) subsets providing an opportunity for parallelization.

The ``invalid guess'' filter mentioned in the discussion of {\em  ELPsolve2}
applies to any ELP solver (and may already be implemented in other solvers).
Yet another idea is to construct a ``hybrid'' solver which runs multiple
different solvers in parallel (e.g., EP-ASP and EHEX), terminating further
computation once any solver completes with the required solution.

% If you want to improve on ELPs, maybe use WVC to get performance
% improvements (eg. yale shooting problem).  Could parallelize Son's
% idea by taking, say 3 bits, then split the search space into 8
% pieces (2^3) and ``hard code'' the constraint on these bits and do
% the rest of Son's.  In general, can parallelize EP-ASP (and
% others). Can combine ideas (e.g., in parallel run EP-ASP and
% EPSsolver.  Generaliziable optimization like ``invalid guesses'',
% things like the WVCs.

%% REFERENCES
\bibliographystyle{acmtrans}
\bibliography{ELPSolverSurvey}

\end{document}